\documentclass[conference]{IEEEtran}
\IEEEoverridecommandlockouts
\usepackage{cite}
\usepackage{amsmath,amssymb,amsfonts}
\usepackage{amsbsy}
\usepackage{mathrsfs}
\usepackage{algorithmic}
\usepackage{graphicx}
\usepackage{textcomp}
\usepackage{xcolor}
\usepackage{bm}
\def\BibTeX{{\rm B\kern-.05em{\sc i\kern-.025em b}\kern-.08em
    T\kern-.1667em\lower.7ex\hbox{E}\kern-.125emX}}
\begin{document}

\title{Test Set Optimization by Machine Learning Algorithms\\

}

\author{\IEEEauthorblockN{1\textsuperscript{st} Kaiming Fu}
\IEEEauthorblockA{\textit{Electrical and Computer Engineering} \\
\textit{University of California, Davis}\\
CA, USA \\
kmfu@ucdavis.edu}
\and
\IEEEauthorblockN{2\textsuperscript{nd} 
Yulu Jin}
\IEEEauthorblockA{\textit{Electrical and Computer Engineering} \\
\textit{University of California, Davis}\\
CA, USA \\
yuljin@ucdavis.edu}
\and
\IEEEauthorblockN{3\textsuperscript{rd}
Zhousheng Chen}
\IEEEauthorblockA{\textit{Electrical and Computer Engineering} \\
\textit{University of California, Davis}\\
CA, USA \\
zhche@ucdavis.edu}
}
\maketitle

\begin{abstract}
Diagnosis results are highly dependent on the volume of test set. To derive the most efficient test set, we propose several machine learning based methods to predict the minimum amount of test data that produces relatively accurate diagnosis. By collecting outputs from failing circuits, the feature matrix and label vector are generated, which involves the inference information of the test termination point. Thus we develop a prediction model to fit the data and determine 
when to terminate testing.
The considered methods include LASSO and Support Vector Machine(SVM) where the relationship between goals(label) and predictors(feature matrix) are considered to be linear in LASSO and nonlinear in SVM. Numerical results show that SVM reaches a diagnosis accuracy of 90.4\% while deducting the volume of test set by 35.24\%.  
\end{abstract}


\section{Introduction}
The process of determining the cause of a failing integrated circuit (IC) is known as failure analysis. The main purpose of this diagnosis is to detect the location of the failure, such as to mitigate the effects based on a specific reconfiguration of the system and to recover from the identified fault with the component’s corrupted state healed or fixed. In these ways, the failure analysis is always operated off-line during normal working time while beginning from the initial assumption. 

The increase cost of this analysis is known as one of the most significant problems in testing a failing chip. The major reason is a large sum of test vectors are needed for large scale circuits test which results in increasing tester memory space, application time, and hence the total cost. The mechanism to find redundant vectors is to create a fault detection table that contains the detection information of all faults detected by each test vector. Various compaction techniques have been proposed by numerous research papers to reduce the cost of test vectors, for example, Brain and Tracy presented a new dictionary organization which concentrate on reducing the size by managing the organization and encoding of the dictionary \cite{chess1999creating}.

However, only a few research works focus on reducing the number of test vectors to reduce the cost of test data storage. The prior work has already shown that a high-volume test data does not mean high accuracy of test output, in other words, the increasing test data volume may not help to improve the diagnosis result\cite{wang2012test}. On the other hand, reducing the test data volume could contribute to a lower cost and allow a reduction in the size of the fault dictionary, which is an important method for fault diagnosis\cite{amati2010formal}.

In this paper, we propose algorithms to reduce diagnostic test vectors for circuits. Linear regression is regarded as one of the most frequently used and well-known statistical models that could be used to solve the problems mentioned above. Regression analysis represents a linear approach to model the relationship between dependent variables and independent variables. By finding the linear model between goals and predictors, it is possible to predict the output values for some specific inputs. Least square is the most important part of the linear regression model which provides the approximation solution for an over-determined system by minimizing the sum of the squares of the residuals \cite{LeastSquare}. In order to find the most reasonable linear relationship impacted by the related explanatory variables, a penalty term is usually added to lower the impact of related variables. 

However, the real relationship between goals and predictors tends to be complex or at least nonlinear. Thus in this work, we use a well-behaved nonlinear model, support vector machines (SVMs) to capture the relationship between circuit testing results and the test termination point. SVM was proposed by Vapnik and his fellows in the 1990s and was applied to analyze data used for classification and regression analysis. Given a set of training examples, a classification model that assigns examples into different categories is generated by mapping the input data into a higher dimensional space and constructing an optimal separating hyper-plane~\cite{suykens1999least}. To perform non-linear classification efficiently, different kernel functions are adopted in the algorithm, which implicitly map their inputs into high-dimensional feature spaces. Hence, in our work, testing data from real circuits are collected and by extracting specific expressive features, the input data matrix of machine learning algorithms is generated. After performing linear regression and SVM models on such input, the termination point is constructed and the performance of such algorithms is discussed. 

\section{Related Work}

Machine learning is integrated into hardware testing by Tang et al. \cite{tang2007}. Since then, it has become more involved in this field. The study of Laura Isabel \cite{Gomez2017} shows the potential of machine learning for gathering useful and useless information in logical analysis and making use of it. A number of related studies of Machine Learning based on hardwares are beginning to emerge. Cheng et al.\cite{Cheng2017} shows that using machine learning method can greatly reduce testing time and cost to accelerate locating systematic defect identification comparing to the physical failure analysis(PFA). Gómez et al.\cite{gomez2016} provides a method to classify different defects. Nelson et al.\cite{nelson2010} and Wang et al.\cite{wang2009} works on volume diagnosis by applying decision forest algorithm to locate bridge defects and stuck-at errors respectively. Moreover, Huang et al.\cite{Huang2018} provides a novel method to find out the meaningful fail log of the circuit which can make industrialized testing more efficient. Similarly, Wang et al.\cite{wang2012test} demonstrated a method to reduce the test volume by applying machine learning techniques including linear regression, k-nearest neighbor, support vector machines, and decision tree. Afterwards, Gomez et al.\cite{Gomez2014} showed that machine learning can not only help address permanent faults, but can also help analysing intermittent faults. 

However, different feature extraction processes have a great impact on the performance of machine learning algorithms. The features used in this work has been shown to obtain relatively high performance in reducing the test volume.  
\section{Proposed Methods}
\subsection{Data Extraction}
For failing circuits, defect responses can be collected, after which the underlying patterns and trends that suggest a termination point for testing are able to be discovered. However, it is difficult to use the collected raw data directly for finding the termination point. Therefore, to create the data matrix X in the fit model, the raw test data is processed and then organized into a set of features that are more readable for a prediction model. 

To be specific, suppose that there are $n$ examples(performed tests), an $n \times 5$ feature matrix $\mathbf{X}$ and an $n \times 1$ response vector $\mathbf{Y}$ are built. Table 1 shows the five features extracted from raw data. We regard the first test case to the first failing test case as the first pattern, the first test case to the second failing test case as the second pattern and etc. The collection of all the possible causes of each failing circuit is defined as a set of intermediate candidates, and define the intersection of all intermediate candidates set as the set of golden candidates. Then define \textit{m} to be  $$\textit{m} = \frac{\text{No.(golden\ candidates)}}{\text{No.(intermediate\ candidates)}},$$ and the minimal element in $m$ is marked as $m_{min}$. The modified element in $\mathbf{Y}$ is defined as: \[  y= \left\{ 
\begin{array}{ll}
      1 & if\ \textit{m}=1, \\
      \frac{m-m_{min}}{1-m_{min}} & if\ \textit{m}\neq1. \\
\end{array} 
\right. \]

\begin{table}[ht]
\centering
\begin{tabular}{|c|c|}
\hline
\textbf{Feature} & \textbf{Feature description}\\
\hline
\textbf{X($\cdot$, 1)} & Amount of inputs to the circuit\\
\hline
\textbf{X($\cdot$, 2)} & Total number of failing patterns have been applied so far\\
\hline
\textbf{X($\cdot$, 3)} & Index of the first failing pattern detected in this circuit\\
\hline
\textbf{X($\cdot$, 4)} & Index of the this failing pattern detected in this circuit\\
\hline
\textbf{X($\cdot$, 5)} & Index of last failing pattern detected in this circuit\\
\hline
\end{tabular}
\label{tab:label}
\caption{Five features extracted from raw data}
\end{table}
\subsection{Linear Regression}

The basic model of linear regression can be expressed as $$\mathbf{Y} = \mathbf{X} \cdot \boldsymbol{\beta} + \boldsymbol{\varepsilon}$$ where $\boldsymbol{\beta}$ is the $5 \times 1$ coefficient vector characterizing the linear model and $\boldsymbol{\varepsilon}$ is the $n \times 1$ residual vector. Specifically, elements in $\boldsymbol{\beta}$ represents the weights of corresponding features in $\mathbf{X}$ to the output $\mathbf{Y}$ and $\boldsymbol{\varepsilon}$ is generated by the  least-square estimation.

To take the model size into consideration, we use LASSO (Least Absolute Shrinkage and Selection) to characterize the relationship between $\mathbf{X}$ and $\mathbf{Y}$, where $\boldsymbol{\varepsilon}$ has been transferred to 
$$\|\mathbf{X} \cdot \boldsymbol{\beta}-\mathbf{Y}\|_{2}^{2}+\alpha\|\boldsymbol{\beta}\|_{2}^{2}$$ and  $\alpha\|\boldsymbol{\beta}\|_{2}^{2}$ is a punish term which can be adjusted by changing the value of tuning parameter $\alpha$.

\subsection{Supporting Vector Machine}


Since the relationship between $\mathbf{X}$ and $\mathbf{Y}$ is not necessarily linear, supporting vector machines is proposed to characterize such complex nonlinear relationship. To solve this problem, the input data is mapped into a higher dimensional space. By applying Kernel functions, nonlinear problem in the lower dimensional space has been transferred into a linear one and thus an optimal separating hyper-plane can be learned \cite{suykens1999least}. In our method, logistic regression SVMs with RBF kernel is applied and the cost function could be expressed explicitly as

\begin{eqnarray}
\min_{\mathbf{\theta}} && \frac{1}{m}\sum_{i=1}^m \left[y^{(i)} (-\log{h_\theta(x^{(i)})}) \right. \nonumber\\
&& \left. +(1-y^{(i)}) (-\log(1-{h_\theta(x^{(i)})})) \right]+\frac{\lambda}{2m}\sum_{i=1}^n \theta_j^2, \nonumber
\end{eqnarray}
where
\begin{eqnarray}
h_\theta = \frac{1}{1 + e^{-\theta^T X}}. \nonumber
\end{eqnarray}

During the training process, the model parameter is learned to minimize the objective function, after which the selected optimal parameter $\theta$ is applied to calculate the prediction label $y$ in the testing process.

\section{Numerical Results}
In this section, we verify that the machine learning algorithms mentioned above can be performed during test process and reduce the test size while maintaining a high defect detection accuracy.

First, we collect test results from $153$ real circuits. Then we calculate the feature matrix $\mathbf{X}$ and label $\mathbf{Y}$, which both contain $5427$ rows. After dividing them into training and testing sets, linear regression and SVM models are performed to fit the data sources $\mathbf{X}$ and $\mathbf{Y}$. The following graphic results are based on the prediction results of the fitted model. 

\begin{figure}[ht]
    \centering
    \includegraphics[scale=0.43]{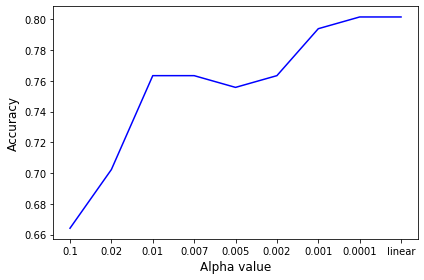}
    \caption{Accuracy at different Values of $\alpha$}
    \label{fig:1}
\end{figure}

Fig.\ref{fig:1} provides the accuracy of prediction results under different punishment terms of LASSO algorithm. The figure shows that the smaller $\alpha$ is, the more accurate the prediction results that locates the certain stuck-at defect will be. Generally, the result of linear regression is better than that of LASSO algorithm.

\begin{figure}[ht]
    \centering
    \includegraphics[scale=0.55]{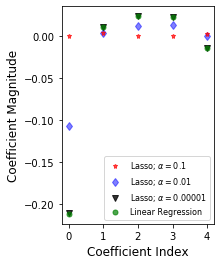}
    \caption{$\beta$ value weights in different $\alpha$ value}
    \label{fig:2}
\end{figure}

Fig.\ref{fig:2} shows how important each feature in \textbf{\textit{X}} is to the prediction results. As we can see, when $\alpha=0.1$, the coordinates of $\mathbf{\beta}$ are all close to $0$ and thus does not fit the data well, which is consistent with the performance result in Fig.~\ref{fig:1}. 

\begin{figure}[ht]
    \centering
    \includegraphics[scale=0.55]{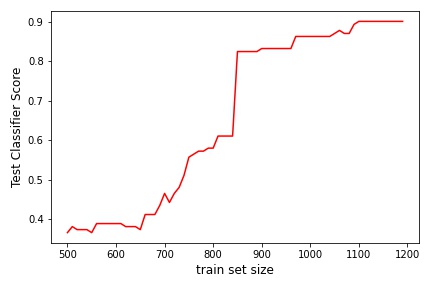}
    \caption{Test classifier score at different train set sizes}
    \label{fig:3}
\end{figure}

Fig.\ref{fig:3} shows how training size affects testing accuracy. For a bigger training set, a higher test classifier score can be achieved, which verifies the accuracy of our algorithms. However, when the size of training set reaches 1080, as the training size increases, the prediction accuracy will not be greatly improved and the highest prediction accuracy is 90.4\%, which is much higher than that of the linear regression model. The reason behind this phenomenon is trivial, that is the data collected from real world can be fitted better by a nonlinear model. Moreover, under the premise of maintaining a diagnosis accuracy of 90.4\%, the volume of test set has been deducted by 35.24\%. 

\section{Summary}
In this paper, the diagnosis process has been optimized by reducing the test volume. Different features from the collected test and response data are extracted and machine learning algorithms are performed to characterize the test termination point. After training, SVM has better diagnosis performance than LASSO and reaches an accuracy of 90.4\% while deducting the volume of test set by 35.24\%.


\bibliographystyle{IEEEtran}
\bibliography{reference}

\end{document}